\documentclass[10pt, a4paper]{article}

\usepackage[final]{lrec2026} 

\usepackage{
url,
comment,
placeins,
booktabs,
multirow,
graphicx,
array,
pgfplots,
xcolor,
pgfplotstable,
caption,
subcaption,
amsmath,
amssymb,
amsfonts,
algorithmic,
textcomp,
lineno,
CJKutf8,
enumitem,
}

\title{Efficient Dialect-Aware Modeling and Conditioning for Low-Resource Taiwanese Hakka Speech Processing}

\name{
\begin{tabular}{c}
An-Ci Peng\textsuperscript{1},
Kuan-Tang Huang\textsuperscript{1},
Tien-Hong Lo\textsuperscript{1}, \\
Hung-Shin Lee\textsuperscript{3},
Hsin-Min Wang\textsuperscript{2},
and Berlin Chen\textsuperscript{1}
\end{tabular} \\
}

\address{
\textsuperscript{1}Dept. Computer Science and Information Engineering, National Taiwan Normal University, Taiwan \\
\textsuperscript{2}Institute of Computer Science, Academia Sinica, Taiwan \\
\textsuperscript{3}United Link Co., Ltd., Taiwan
}

\abstract{
Taiwanese Hakka is a low-resource, endangered language that poses significant challenges for automatic speech recognition (ASR), including high dialectal variability and the presence of two distinct writing systems (Hanzi and Pinyin).
Traditional ASR models often encounter difficulties in this context, as they tend to conflate essential linguistic content with dialect-specific variations across both phonological and lexical dimensions.
To address these challenges, we propose a unified framework grounded in the Recurrent Neural Network Transducers (RNN-T).
Central to our approach is the introduction of dialect-aware modeling strategies designed to disentangle dialectal ``style'' from linguistic ``content'', which enhances the model's capacity to learn robust and generalized representations.
Additionally, the framework employs parameter-efficient prediction networks to concurrently model ASR (Hanzi and Pinyin).
We demonstrate that these tasks create a powerful synergy, wherein the cross-script objective serves as a mutual regularizer to improve the primary ASR tasks.
Experiments conducted on the HAT corpus reveal that our model achieves 57.00\% and 40.41\% relative error rate reduction on Hanzi and Pinyin ASR, respectively.
To our knowledge, this is the first systematic investigation into the impact of Hakka dialectal variations on ASR and the first single model capable of jointly addressing these tasks.\\
\newline 
}

\pgfplotsset{compat=1.18}

\begin{document}
\begin{CJK*}{UTF8}{bsmi}
\maketitleabstract

\section{Introduction}

Automatic Speech Recognition (ASR) has undergone significant advancements, transitioning from traditional hybrid models to advanced end-to-end (E2E) architectures.
Key innovations in this space include Connectionist Temporal Classification (CTC) \cite{garves2006}, Attention-based Encoder-Decoder (AED) \cite{vaswani2017}, and Recurrent Neural Network Transducers (RNN-T) \cite{graves2012}.

Among these E2E paradigms, the RNN-T model has garnered significant attention.
It offers a robust and flexible framework that achieves an optimal balance between high recognition accuracy and the low-latency streaming capabilities essential for real-time applications \cite{he2019, sainath2020}, thereby serving as the foundation of our study.
Nonetheless, while the architecture of RNN-T has demonstrated strong performance in widely spoken languages such as English and Mandarin \cite{wang2024,zhang2022}, its application to low-resource and dialectally diverse languages continues to pose substantial challenges.

Taiwanese Hakka, once the second most prominent language in Taiwan \cite{lai2023}, is now classified as endangered due to the predominance of Mandarin Chinese.
The development of robust ASR tools for Taiwanese Hakka are essential for language preservation, cultural archiving, and enhancing media accessibility.
However, the construction of such tools encounters two significant challenges.
First, the language faces a substantial data scarcity issue: it is not only low-resource but also showcases considerable dialectal variability \cite{lai2001,chang2024}, complicating the creation of a unified, generalized model from the limited available data.
Second, practical applications in educational and linguistic contexts necessitate support for two distinct writing systems: Taiwanese Hakka Hanzi \cite{moe2009,moe2010} (Hanzi, 漢字), which utilizes traditional Chinese characters, and the Hakka Romanization System \cite{moe2012,moe2024} (Pinyin, 拼音).

Although recent approaches have sought to tackle dialectal variability \cite{li2024}, many strategies involve adapting specific model components.
These approaches typically encompass the training of distinct prediction networks for each dialect, which are subsequently integrated during the post-training phase \cite{fukuda2022}, or conditioning the decoder on an explicit dialect ID token \cite{shiota2022}.
These methods, however, are often evaluated on relatively high-resource languages and may not be directly applicable to low-resource scenarios where data is insufficient to train robust, specialized components.
Moreover, by concentrating primarily on the prediction network, these approaches largely neglect the fundamental challenge of representation conflation within the shared encoder.

When trained on limited and diverse datasets, a standard encoder is compelled to conflate core linguistic content with dialect-specific variations, affecting both phonological and lexical representations.
This conflation poses significant challenges: it necessitates that the RNN-T's prediction network, which serves as its internal language model, learns multiple, redundant mappings for the same linguistic concepts in an inefficient manner.

To address this issue, and as a central contribution of our work, we introduce a dialect integration module consisting of two complementary components\footnote{Code: \url{https://anonymous.4open.science/r/hakka_exp-664D}}.
The first part of this module is a dialect-aware modeling strategy.
This strategy aims to explicitly condition the model on dialect identity \cite{punjabi2021}, guiding the encoder to capture dialect-specific acoustic and phonological features.
The second part of the dialect integration module is dialect conditioning.
This method enables the prediction network to function as a dialect-conditional language model, dynamically adapting its predictions---such as word choice and phonetic-to-character mappings---according to the active dialect, rather than attempting to model all dialects using a single, compromised set of parameters.
This targeted conditioning facilitates effective generalization by the model, even in scenarios characterized by limited training data.

The second challenge of supporting two writing systems is addressed by a unified multi-task learning (MTL) framework \cite{wang2023,wang2021,hussein2025}.
Rather than training separate, specialized models for each task, our unified approach extends a single RNN-T architecture with parameter-efficient prediction networks \cite{ghodsi2020}, enabling the joint generation of Hanzi and Pinyin without a prohibitive increase in model parameters.
These two ASR tasks are complementary: the phonetically-grounded Pinyin task compels the model to maintain acoustically precise representations, while the semantically rich Hanzi task simultaneously requires it to capture high-level linguistic context to resolve ambiguities.
This synergy creates a complementary objective that regularizes the shared encoder, forcing it to learn a representation that is both acoustically precise and linguistically context-aware.

Our experiments conducted on the HAT corpus \cite{lai2023} validate that our single, unified model achieves significant improvements over all single-task baselines, resulting in 57.00\% and 40.41\% relative error rate reductions on the Hanzi and Pinyin ASR tasks, respectively.
The main contributions of this work are summarized as follows: 
\begin{enumerate}[noitemsep,leftmargin=*]
\item \textbf{Systematic Analysis of Dialectal Variation:}
To our knowledge, this is the first systematic study of Hakka dialectal variation's impact on ASR.
We propose a novel modeling strategy that explicitly conditions the RNN-T model on dialect identity.
This allows the prediction network to function as a dialect-conditional language model, dynamically adapting its output.
The result is a more robust and generalized model that provides a path forward for other low-resource, dialect-rich languages.
\item \textbf{A Unified Framework for Multi-Orthography Speech Processing:}
We present a single, unified framework to jointly perform ASR (in Hanzi and Pinyin) using Taiwanese Hakka as a case study.
We demonstrate that this architecture creates a powerful synergy, where the cross-lingual and cross-script objectives mutually regularize the model.
This provides a parameter-efficient blueprint for handling similar challenges in other multi-orthography languages.
\end{enumerate}

\section{Background}

This section provides the foundational context for our research.
We begin by exploring the core architecture of the Recurrent Neural Network Transducer (RNN-T), which serves as the basis for our proposed model.
Following this, we delve into the linguistic intricacies of Taiwanese Hakka, highlighting its two primary writing systems, Hanzi and Pinyin, as well as the main dialects relevant to this study.

\subsection{Neural Transducer (RNN-T)}
\label{sec:rnn_t}
The neural transducer network, commonly known as the Recurrent Neural Network Transducer (RNN-T) \cite{graves2012}, is an end-to-end architecture specifically tailored for sequence-to-sequence tasks.
The architecture typically comprises three interrelated components: an Encoder Network, a Prediction Network, and a Joint Network.

The process initiates with the encoder network, $\text{Enc}(\cdot)$.
At the current time step $t$ in the input acoustic feature sequence $\mathbf{x}_{1:T}$ (where $T$ denotes the length of the input audio features), the encoder transforms the frame $\mathbf{x}_{t}$ into a high-level acoustic representation $\mathbf{h}^{\text{Enc}}_{t}$:
\begin{equation}
\mathbf{h}^{\text{Enc}}_{t} = \text{Enc}(\textbf{x}_{t}).
\label{eq:encoder}
\end{equation}

Simultaneously, the prediction network $\text{Pred}(\cdot)$ operates in an autoregressive manner, leveraging the previously predicted token sequence $\hat{\mathbf{y}}_{1:u}$ to generate a hidden representation $\mathbf{h}^{\text{Pred}}_{u+1}$ at the $(u{+}1)$-th output step. This network serves as an internal language model:
\begin{equation}
\mathbf{h}^{\text{Pred}}_{u+1} = \text{Pred}(\hat{\mathbf{y}}_{1:u}).
\label{eq:prediction}
\end{equation}

Finally, the joint network, $\text{Joint}(\cdot)$, integrates the outputs $\mathbf{h}^{\text{Enc}}_t$ and $\mathbf{h}^{\text{Pred}}_{u+1}$ from the encoder and the prediction network through a feed-forward layer.
This network predicts the posterior probability distribution over the vocabulary:
\begin{equation}
P(\hat{y}_{u+1}|x_t, \hat{\textbf{y}}_u) = \sigma(\text{Joint}(\mathbf{h}^{\text{Enc}}_t, \mathbf{h}^{\text{Pred}}_{u+1}))
\label{eq:joint}
\end{equation}

While the training objective is defined by maximizing the probability of the correct output sequence $\mathbf{y}$ given the input sequence $\mathbf{x}$ across all possible alignments between the input steps $t$ and output steps $u$, this is computed efficiently using dynamic programming.
This joint optimization enables the RNN-T to concurrently learn both robust acoustic and linguistic contexts within a unified differentiable framework.

In recent years, substantial research has focused on enhancing the prediction network within RNN-T architectures \cite{zhan2022}.
A study demonstrates that a stateless variant (RNNT-SLP) exhibits performance comparable to the original recurrent version \cite{ghodsi2020}.
The RNNT-SLP architecture relies solely on the most recent output tokens(the last one or the last two), achieving comparable recognition accuracy while substantially reducing model parameters and architectural complexity.
These findings suggest that the prediction network primarily serves the function of alignment rather than language modeling, thereby motivating the development of more compact and parameter-efficient RNN-T designs suitable for low-resource ASR.

\begin{figure*}
\centering
\includegraphics[width=1\linewidth]{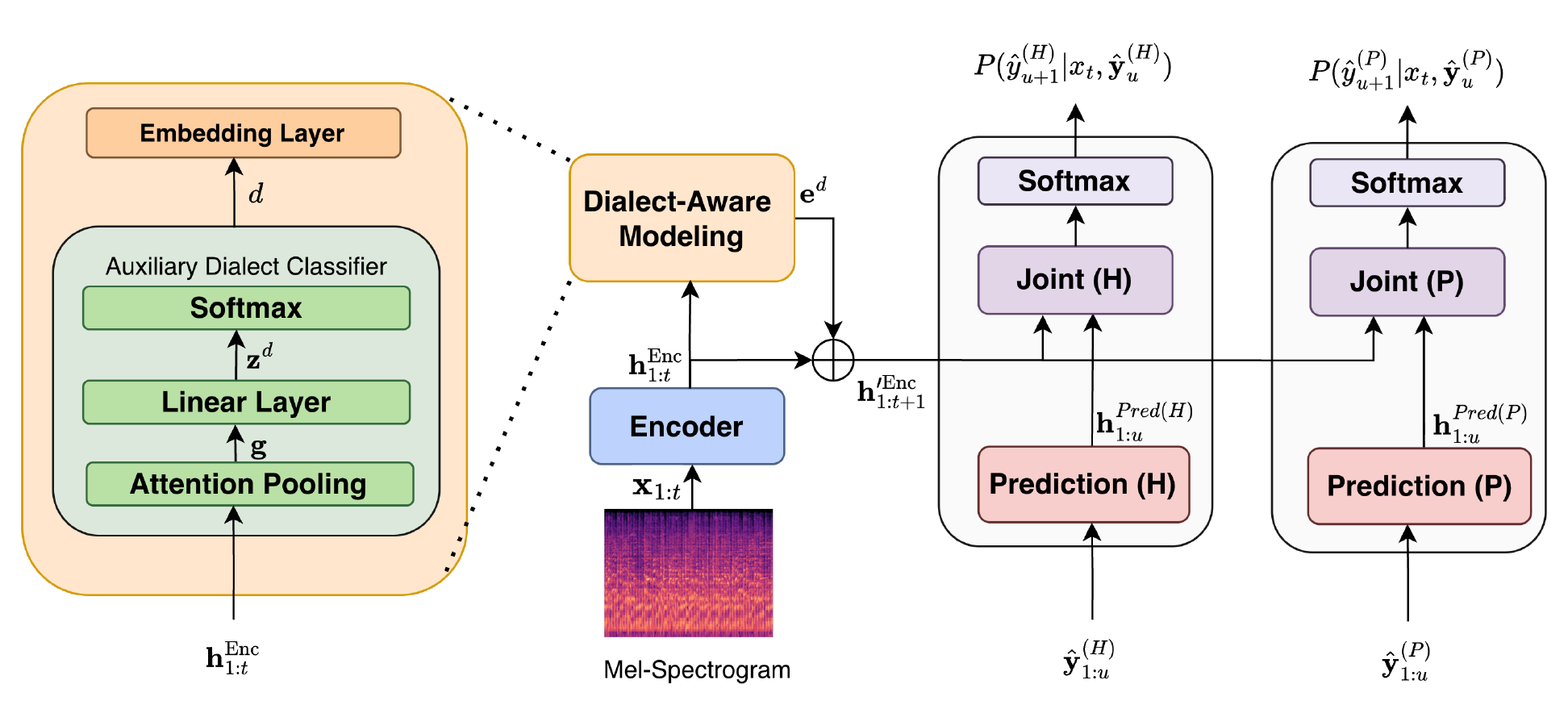}
\caption{Overall architecture of the proposed method, including an RNN-T architecture with multiple joint networks and prediction networks for Hanzi(H) and Pinyin(P) ASR task, and a dialect-aware module.}
\label{fig:archi_all}
\end{figure*}

\subsection{Taiwanese Hakka}

The complexity of the Taiwanese Hakka language is rooted in its dialectal variation and its dual-system orthography.
Dialect differences manifest in tone realization, phonetics, and lexical usage.
Taiwanese Hakka is a language group form of Hakka dialects, such as Sixian (四縣), NanSixian (南四縣), Hailu (海陸), Dapu (大埔), Raoping (饒平), and Zhaoan (詔安).
Due to dataset limitations, this study focuses on three dialects, Sixian, Hailu, and NanSixian.

For the promotion, education and preservation of the language, Taiwanese Hakka utilizes two officially standardized writing systems proposed by the Ministry of Education (MOE) of Taiwan:

\textbf{Taiwanese Hakka Hanzi (Hanzi)}, standardized by the MOE between 2009 and 2010 \cite{moe2009, moe2010}, employs traditional Chinese characters adapted to represent Hakka-specific lexicon and phonetic nuances.
The Hanzi form emphasizes the morpho-semantic aspect of the language, encoding Hakka phonological distinctions through character choice and usage conventions.

\textbf{Taiwanese Hakka Romanization System (Pinyin)}, introduced by the MOE in 2012, uses the Latin alphabet for syllable representation, and the tonemes (tones) are represented using Arabic numerals based on the relative pitch variation of the tones \cite{moe2012,moe2024}.
Pinyin serves as the phonetic transcription standard, closely following the acoustic realization of the spoken language.
These two writing forms form the dual targets for our speech recognition tasks, while the inherent differences among the three core dialects (Sixian, NanSixian, and Hailu) serve as the foundation for our dialect-aware modeling experiments.
Table \ref{tab:hakka_dialect_example} shows a writing example of the Hanzi, Pinyin across dialects like Sixian, Nansixian, and Hailu, with the same meaning.

\begin{table}[!t]
\centering
\setlength{\tabcolsep}{6pt}
\renewcommand{\arraystretch}{1.2}
\begin{tabular}{lll}
\toprule
\textbf{Dialect} & \textbf{Hanzi} & \textbf{Pinyin} \\
\midrule
Sixian & 佢發譴哩 & gi11 fad2 kien31 le24 \\
Nansixian & 佢發譴吔 & i11 fad2 kien31 e24\\
Hailu & 佢發閼咧 & gi55 bod5 ad5 le53 \\
\bottomrule
\end{tabular}
\caption{Example sentences expressing the same meaning, ``He is angry'', across three Taiwanese Hakka dialects. The corresponding Mandarin translation is ``他生氣了''.
Notably, the Hanzi forms differ in characters such as ``譴'', ``閼'', and ``吔'', reflecting dialect-specific lexical variation. The Pinyin representations also exhibit distinct syllabic realizations, for instance ``kien31'' versus ``ad5''. Moreover, even when the same Hanzi character is shared across dialects, the corresponding pronunciations often diverge, highlighting the phonetic variation inherent to Taiwanese Hakka.}
\label{tab:hakka_dialect_example}
\end{table}

\section{Methodology}

The overall architecture of our proposed method is illustrated in Figure \ref{fig:archi_all}.
The system is built upon a multi-task RNN-T framework, comprising three primary components: 1) a shared acoustic encoder; 2) two distinct decoders, each equipped with its dedicated prediction and joint network, tailored for the Hanzi (H) and Pinyin (P) tasks; and 3) a novel dialect-aware module.
This dialect-aware module generates a dialect embedding ($\mathbf{e}^d$) that is fused with the shared encoder's hidden representation ($\mathbf{h}^{\text{Enc}}$).
This integration results in a dialect-conditioned representation that is subsequently input into both ASR decoders, allowing them to adjust their predictions effectively.

In the following subsections, we will first describe the architecture of the shared-encoder multi-task RNN-T backbone (Section~\ref{sec:multi-task-learning}).
We will then detail the components of our dialect-aware module, which includes the Auxiliary Dialect Classifier (ADC) and Dialect Information Integration (DII) (Section~\ref{sec:dialect_aware}).
Finally, we will present several dialect conditioning strategies that adjust the target sequences (Section~\ref{sec:dialect_conditioning}).

\subsection{Multi-task Learning}\label{sec:multi-task-learning}

We adopt a shared-encoder multi-task RNN-T architecture to simultaneously handle the two required output tasks: Hakka Hanzi recognition and Hakka Pinyin recognition.
As illustrated in Figure \ref{fig:archi_all}, the architecture consists of a single shared encoder and separate, task-specific prediction networks and joint networks for each output.

The shared encoder transforms the input audio feature sequence $\mathbf{x}_{1:T}$ of length $T$ into a hidden representation $\textbf{h}_{1:T}^{\text{Enc}}\in \mathbb{R}^{T \times D}$, where $T$ denotes the length of the hidden representation and $D$ indicates the dimension of the encoder.
This representation is subsequently distributed to two distinct branches, associated with the Hanzi and Pinyin linguistic tasks.
The overall ASR loss ($\mathcal{L}_{\text{ASR}}$) is computed by aggregating the individual pruned RNN-T losses \cite{kuang2022} from both the Hanzi and Pinyin tasks, thereby enhancing the time and memory efficiency during training.

\subsection{Dialect-Aware Modeling}
\label{sec:dialect_aware}

Here we introduce the first component of our dialect integration module: dialect-aware modeling.
In the RNN-T architecture, the encoder produces the acoustic representation, significantly impacting the accuracy of the final token sequence prediction.
To enhance the robustness of this representation by incorporating explicit dialect knowledge, we investigate two strategies for infusing dialect information into the features generated by the encoder: the first is the adoption of an Auxiliary Dialect Classifier (ADC), and the second is Dialect Information Integration (DII). 

\subsubsection{ADC: Auxiliary Dialect Classifier}
In the ADC strategy, we integrate an auxiliary classification head that is connected to the output of the encoder $\textbf{h}^{\text{Enc}} \in \mathbb{R}^{T \times D}$.
First, this head aggregates the variable-length sequence $\textbf{h}^{\text{Enc}}$ into a single, fixed-dimensional vector $\mathbf{g} \in \mathbb{R}^D$ through the application of attention pooling.
Subsequently, this vector $\mathbf{g}$ is forwarded through a linear layer to generate the dialect logits $\mathbf{z}^d$.

During training, the auxiliary loss ($\mathcal{L}_{\text{A}}$) is computed using the Cross-Entropy (CE) loss with the softmax output of the logits, $\sigma(\mathbf{z}^d)$, defined as follows:
\begin{equation}
\mathcal{L}_A = \mathcal{L}_{\text{CE}}(\sigma(\mathbf{z}^d), d) ,
\label{eq:loss-A}
\end{equation}
where $\sigma$ denotes the softmax function, and $d$ corresponds to the ground-truth dialect label.

The final objective $\mathcal{L}_\text{Final}$ is formulated as a weighted sum of the primary RNN-T loss ($\mathcal{L}_{\text{ASR}}$) and the auxiliary loss, appropriately balanced by a hyperparameter $\lambda$:
\begin{equation}
\mathcal{L}_{\text{Final}} = \mathcal{L}_{\text{ASR}} + \lambda \cdot \mathcal{L}_A .
\label{eq:loss-ADC}
\end{equation}

\subsubsection{DII: Dialect Information Integration}
The DII strategy explicitly conditions the entire sequence transduction process on the known dialect identity.
This methodology considers the dialect label as a crucial linguistic feature, necessitating the joiner network to generate output tokens informed by both acoustic features and the specific dialect context.

We first map the discrete dialect ID (e.g., NanSixian, Sixian, Hailu) of an audio segment to a continuous dialect embedding $\mathbf{e}^{d} \in \mathbb{R}^{D}$ through an embedding layer.
During training, the ground-truth dialect label is used to generate this embedding (i.e., employing teacher-forcing).
In contrast, during inference, we leverage the dialect ID predicted by the ADC module to retrieve the corresponding embedding.

Specifically, we employ a prefix conditioning approach whereby the dialect embedding $\mathbf{e}^{d}$ is concatenated to the acoustic hidden representation $\mathbf{h}^{\text{Enc}}$, thereby yielding the augmented representation $\mathbf{h}'^{\text{Enc}}$:
\begin{equation}
\mathbf{h}'^{\text{Enc}} = [\mathbf{e}^{d}; \mathbf{h}^{\text{Enc}}] , \quad \mathbf{h}'^{\text{Enc}} \in \mathbb{R}^{(T+1) \times (D)}.
\end{equation}

This augmented sequence $\mathbf{h}'^{\text{Enc}}$ is then fed into the joint networks.
By integrating dialect information at the input stage of joint prediction, the model is encouraged to refine its token prediction probabilities according to the respective dialect, thereby enhancing recognition accuracy for that specific dialectal variant.

\subsection{Dialect Conditioning}
\label{sec:dialect_conditioning}

To enhance the model's performance in a multi-dialect scenario, we explore a set of strategies of integrating dialect information by adding the dialect index (denoted as $d$) directly to the prediction target sequence $\mathbf{y} = (y_1, \dots, y_U)$ of each decoder, where $U$ is the length of the target sequence.
To ensure the dialect index exists within the same vocabulary space for all target sequences (Hanzi and Pinyin), we expand the token table of each decoder by a size of $K$, where $K$ is the total number of dialect classes.
The last $K$ tokens in each expanded token table are reserved to represent the corresponding $d$ for that output target.
For instance, if the size of the original Hanzi token table is $V_{\text{Hanzi}} = 500$, and we have $K=3$ dialects (NanSixian, Sixian, and Hailu), the table size is extended to $V_{\text{Hanzi}} + 3 = 503$.
In this extended vocabulary, tokens 500, 501, and 502 explicitly represent the NanSixian, Sixian, and Hailu dialects, respectively.
This mechanism is replicated across each ASR task, allowing the decoders to predict not only the linguistic sequence but also the dialectal context simultaneously.
These strategies provide an explicit dialect signal that guides the model's predictions based on the source speech's dialect.
Concrete examples of these three strategies are illustrated in Table \ref{tab:hakka_strategy_example}.

\begin{table}[t]
\centering
\setlength{\tabcolsep}{23pt} 
\renewcommand{\arraystretch}{1.2} 
\begin{tabular}{ll}
\toprule
\textbf{Strategy} & \textbf{Target Sequence} \\
\midrule
PSC & 佢發閼咧$d$ \\
PRSC & $d$佢發閼咧\\
TIC & 佢$d$發$d$閼$d$咧$d$\\
\bottomrule
\end{tabular}
\caption{Illustration of example target sequence in each dialect conditioning strategy.
$d$ means the dialect ID.}
\label{tab:hakka_strategy_example}
\end{table}

\textbf{Post-Sequence Conditioning (PSC).} This strategy appends the dialect ID ($d$) to the end of the original target sequence.
The core idea behind this placement is to provide an explicit signal to the Joint Network, ensuring the dialectal factor is integrated into the loss calculation for the final token prediction.
This configuration compels the decoder to first generate the complete linguistic content.
The conditioned target sequence processed by this strategy is $\textbf{y}_\text{PSC} = (y_1, \dots, y_U, d)$.

\textbf{Pre-Sequence Conditioning (PRSC).}
In this approach, the dialect ID is prefixed to the beginning of the target sequence.
This formulation equips the decoder with essential dialectal context prior to initiating the generation of linguistic tokens, thereby serving as an initial prompt or conditioning cue.
The conditioned target sequence is $\textbf{y}_\text{PRSC} = (d, y_1, \dots, y_U)$.

\textbf{Token-Interleaved Conditioning (TIC).}
The proposed strategy interleaves the dialect ID within the output sequence.
It incorporates the dialect ID by following the processing of each linguistic token.
This dense, token-level integration guarantees that the decoder's prediction for each subsequent token is continuously reinforced and conditioned by the corresponding target dialect.
The conditioned target sequence is $\textbf{y}_\text{TIC} = (y_1, d, y_2, d, \dots, y_U, d)$.

\section{Experimental Setup}
\subsection{Datasets}

We evaluated the strategies mentioned in this paper on the Hakka Across Taiwan (HAT) corpus \cite{lai2023}.
The HAT corpus is particularly valuable as it contains speech data spanning three major Taiwanese Hakka dialects: Hailu, Sixian, and NanSixian.
To simulate a low-resource scenario and evaluate model robustness across these distinct linguistic varieties, we constructed a dialect-balanced subset for our experiments.

From the corpus, we selected 60 speakers, 20 from each dialect, and partitioned this subset into speaker-disjoint training, development, and test sets.
The training set contains data from 18 speakers per dialect, resulting in a total of 54 speakers.
The development and test sets are balanced across dialects, each containing 3 speakers, one from each dialect.
This partitioning ensures that all dialects are adequately represented in the training, development, and test data.
Detailed statistics of this dataset, including duration and utterance counts for each split, are presented in Table \ref{tab:corpus}.

\begin{table}[t]
\centering
\setlength{\tabcolsep}{3.5pt}
\begin{tabular}{lcc}
\toprule
\textbf{Split} & \textbf{Duration (hours)} & \textbf{\# Utterances} \\
\midrule
Train & 73.91 & 36,867 \\
Development & 3.95 & 2,330 \\
Test & 4.46 & 2,439\\
\midrule
Total & 82.32 & 41,636 \\
\bottomrule
\end{tabular}
\vspace{-5pt}
\caption{Statistics of the dialect-balanced subset constructed from the HAT corpus, including total duration and number of utterances in each split.}
\label{tab:corpus}
\vspace{-5pt}
\end{table}

\subsection{Implementation Details}

The encoder employs a Zipformer \cite{yao2024} architecture.
A Zipformer block, which is approximately twice as deep as a Conformer block \cite{gulati2020}, incorporates self-attention, convolution, and feed-forward modules.
Our encoder consists of 6 Zipformer stacking blocks.
The number of attention heads for each block is set to $\{4,4,4,8,4,4\}$, while the convolution kernel sizes are set to $\{31,31,15,15,15,31\}$.
The feed-forward dimensions per block are specified as $\{512, 768, 1024, 1536, 1024, 768\}$.
For the prediction network, we adopt the RNN-T-SLP architecture \cite{ghodsi2020}.

Models are optimized using the ScaledAdam optimizer \cite{yao2024} with a learning rate set to 0.01.
The parameter $\lambda$ in Eq. \ref{eq:loss-ADC} is set to 0.5, which is tuned on the development set.
All models undergo training for a total of 30 epochs.

\subsection{Evaluation Metrics}
We evaluate model performance using metrics specifically tailored for each task.
For the Hakka ASR tasks (Hanzi and Pinyin), we use character error rate (CER) and syllable error rate (SER).
CER quantifies the text fidelity at the character level, defined as: $\text{CER} = (S + D + I) / N_{\text{char}}$, where $S$, $D$, and $I$ represent the counts of substitutions, deletions, and insertions, respectively; $N_{\text{char}}$ is the total number of characters in the reference.
The calculation principle for SER mirrors that of CER; however, the unit of measure is based on syllables.

\section{Results and Discussion}\label{result}

Before detailing our main findings, we first validate our foundational choice of using a mixed-dialect training set.
Mixing data from multiple dialects allows the model to learn dialect-invariant acoustic and linguistic patterns.
As shown in Table \ref{tab:hakka_mix_dialect_eff}, training on the ``Mix'' set (6.07\% in CER) significantly outperforms training on any single dialect (e.g., Hailu at 17.27\% in CER).
This demonstrates the effectiveness of mixed-dialect training in enhancing performance.
Therefore, all subsequent experiments in this paper are conducted using this mixed-data configuration.

With this foundational result established, we conduct an analysis across three main aspects of the proposed framework.
First, we examine the synergistic effect of multi-task learning, evaluating how jointly modeling multiple linguistic representations (Hanzi and Pinyin) enhances recognition performance across tasks.
Next, we analyze the effectiveness of integrating dialect information, demonstrating how dialect-aware and dialect-conditioning strategies contribute to improved modeling and generalization in multi-dialect settings.
Finally, we examine the parameter and time efficiency of our models, illustrating that the proposed multi-task and dialect-aware architecture achieves notable accuracy gains with only modest increases in model size and minimal impact on inference speed.

\begin{table}[t]
\centering
\setlength{\tabcolsep}{6.5pt}
\renewcommand{\arraystretch}{1.2}
\begin{tabular}{lcc}
\toprule
\textbf{Dialect} & \textbf{Hanzi} \small(CER \%) & \textbf{Pinyin} \small(SER \%)\\
\midrule
Sixian & 10.61 & 8.56 \\
NanSixian & 8.35 & 10.60\\
Hailu & 17.27 & 11.25\\
\midrule
Mix & \textbf{6.07} & \textbf{8.29}\\
\bottomrule
\end{tabular}
\caption{Comparison of Hanzi CER and Pinyin SER for single-task models trained on individual dialects. ``Mix'' denotes training with data combined from all three dialects, resulting in the best overall performance across both tasks.}
\label{tab:hakka_mix_dialect_eff}
\end{table}

\subsection{Effect of Multi-Task Learning}

Here we analyze the benefits of multi-task learning (MTL) as summarized in Table \ref{tab:ablation_dialect_strategies}.
In comparison to the single-task baseline models, the $\mathit{MTL}_{HP}$ architecture demonstrates significant enhancements across both tasks.
Specifically, the $\mathit{MTL}_{HP}$ model obtains relative improvements of 8.23\% in Hanzi CER and 3.61\% in Pinyin SER.
This indicates a distinct synergistic effect, suggesting that these two ASR tasks are complementary.
The phonetically-grounded Pinyin task compels the model to maintain acoustically precise representations, while the semantically rich Hanzi task simultaneously requires it to capture high-level linguistic context to resolve ambiguities.
By training on both, the shared encoder is regularized to learn a more robust representation that benefits both tasks.

\begin{table*}[t]
\centering
\setlength{\tabcolsep}{12.7pt}
\begin{tabular}{lcccccl} 
\toprule
\textbf{Method} & \multicolumn{2}{c}{\textbf{Hanzi}} & \multicolumn{2}{c}{\textbf{Pinyin}} & \multicolumn{2}{c}{\textbf{Dialect}}\\ 
\cmidrule(lr){2-3}
\cmidrule(lr){4-5}
\cmidrule(lr){6-7}
Eval. & \small CER \% & \small Rel. \% & \small SER \% & \small Rel. \% & \multicolumn{2}{c}{\small Acc. \%}\\ 
\midrule
$\mathit{VAN}_{H}$ & 6.07& -- & -- & -- & \multicolumn{2}{c}{--} \\
$\mathit{VAN}_{P}$ & -- & -- & 8.29 & -- & \multicolumn{2}{c}{--} \\
$\mathit{MTL}_{HP}$ & 5.57& 8.23 & 7.99 & 3.61 & \multicolumn{2}{c}{--}\\
\midrule
\textbf{Dialect-Aware Modeling} & & \\
\, w/ ADC & 5.29 & 12.80 & 7.85 & 5.31 & \multicolumn{2}{c}{\textbf{100.00}}\\
\, w/ DII& 5.21 & 14.16 & 7.70 & 7.16 & \multicolumn{2}{c}{--}\\
\, w/ ADC and DII & 5.32 & 12.35 & 7.84 & 5.43 & \multicolumn{2}{c}{99.96} \\
\midrule
\textbf{Dialect Conditioning} & & \\
\, w/ PSC & 5.41 & 10.87 & 7.76& 6.39 & \multicolumn{2}{c}{--}\\
\, w/ PRSC & 10.54 & -73.64 & 12.88& -55.36 & 88.88 & 88.80\\
\, w/ TIC & 2.71 & 55.35 &4.98& 39.90 & 88.76 & 88.76\\
\, w/ ADC , DII and TIC & \textbf{2.61} & 57.00 & \textbf{4.97} & 40.41 & \multicolumn{2}{c}{\textbf{100.00}} \\
\bottomrule
\end{tabular}
\caption{Comparison of CER of Hanzi and SER of Pinyin of the model with each dialect integration strategy.
$\mathit{VAN}$ (Vanilla) denotes models trained on a single task, while $\mathit{MTL}$ denotes joint training on multiple tasks.
H and P refer to the Hanzi, Pinyin tasks, respectively.
Rel.\% indicates relative improvement over the corresponding $\mathit{VAN}$ baseline.
All strategies listed under "Dialect-Aware Modeling" and "Dialect Conditioning" are built upon the $\mathit{MTL}_{HP}$ model.
}
\label{tab:ablation_dialect_strategies}
\end{table*}

\subsection{Effectiveness of Integrating Dialect}
\subsubsection{Dialect-aware Modeling Results}
We evaluated the two encoder-side dialect-aware methods, DII and ADC.
As shown in Table \ref{tab:ablation_dialect_strategies}, DII proved to be more effective in reducing the CER to 5.21\% and SER to 7.70\%.
Due to its architectural design as a conditioning module, DII itself cannot predict dialects.
Unlike methods that manually integrate dialect information, ADC can predict the dialect of each utterance.
While DII achieves a slightly lower error rate (5.21\% vs. 5.29\%) than ADC, it requires ground-truth IDs during inference. 
Additionally, the ADC module predicts dialects with near-perfect accuracy (100\%) and regularizes hidden states to extract discriminative features, making it essential for real-world ASR when dialect labels are unavailable.

\subsubsection{Dialect Conditioning Results}
The PSC strategy provided only marginal gains over the baseline (Hanzi CER 5.41\%, Pinyin SER 7.76\%), indicating that appending a dialect token offers limited benefit.
However, the model consistently failed to output the dialect ID during inference, likely due to the RNN-T decoder's lack of an explicit boundary cue to trigger the final token.

In contrast, the PRSC strategy, which places the dialect ID at the beginning of the sequence, severely degraded ASR performance (Hanzi CER 10.54\%, Pinyin SER 12.88\%) despite achieving high dialect accuracy (88.88\%).
This suggests that enforcing a fixed initial dialect cue strongly biases the prediction network's state, impeding its ability to generate the initial linguistic tokens.

The TIC strategy stands out as a strongly effective conditioning method, achieving a dramatic performance leap (CER=2.71\%, SER=4.98\%). 
This demonstrates a key finding: providing a continuous dialect context directly to the target prediction network is more critical for reducing recognition errors in a multi-dialect scenario.
The dense, temporal repetition of the dialect ID throughout the sequence enhances the model stability and robustness for a multi-dialect scenario.

\begin{figure}
\centering
\includegraphics[width=1.0\linewidth]{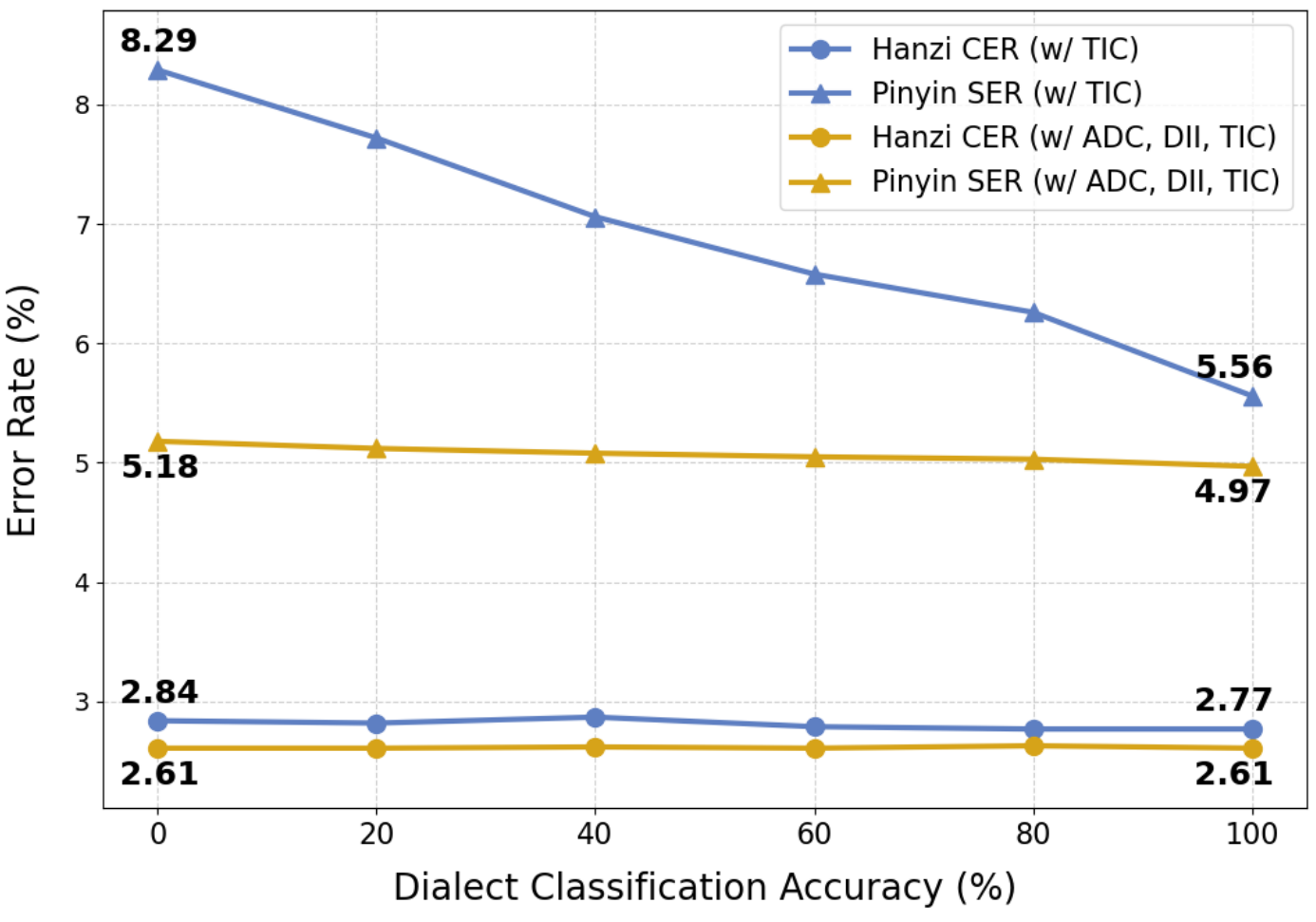}
\caption{Effect of dialect Acc on ASR performance of our method on the test set. 
Performance is measured using CER (\%) for Hanzi and SER (\%) for Pinyin.}
\label{fig:tic_lines}
\end{figure}

To investigate the model's reliance on TIC, we designed a stress test. During inference, instead of using the model's self-predicted dialect tokens, we force-fed the decoder with dialect IDs of controlled correctness. We varied this correctness from 100\% (always the correct ground-truth ID) down to 0\% (always an incorrect dialect ID) and measured the impact on ASR performance.
As shown in Figure \ref{fig:tic_lines}, the model trained with TIC achieved baseline-level performance on Pinyin when the provided dialect IDs were entirely incorrect (0\%), while Hanzi still outperformed the baseline.
Performance gradually improved as the correctness increased, suggesting that accurate dialect cues can effectively guide decoding toward dialect-consistent linguistic forms.

The model trained with ADC, DII, and TIC consistently outperformed the TIC-only system across all evaluated correctness levels.
This finding indicates that the encoder-side conditioning (via ADC and DII) provides a more robust and comprehensive dialect-aware acoustic representation, rather than being redundant to the decoder-side TIC.
By compelling the encoder to capture dialect-specific phonological patterns directly from the audio, these modules create a more stable foundation for the ASR task.
This enhancement makes the overall model more resilient, allowing it to maintain superior recognition performance even when the dialect context provided to the decoder is imperfect or noisy.

\subsection{Complexity and Efficiency Analysis}
Table \ref{tab:params_and_rtf} compares the model complexity and runtime efficiency between the single-task baseline and the proposed multi-task model enhanced with ADC, DII, and TIC strategies.
Despite incorporating multiple dialect-aware and multi-task modules, the proposed model increases the total parameter count only moderately, from 71.73 M to 78.71 M. 
Remarkably, the multi-task model achieves a higher RTFx (Inverse Real-Time Factor), indicating that improved acoustic representations can facilitate more efficient decoding. 
The decoding complexity remains unchanged, as no additional search operations are introduced, and the computational overhead of the auxiliary components is negligible compared to the encoder-decoder backbone. 
These results demonstrate that our design not only enhances modeling capacity and multi-dialect adaptability but also slightly improves real-time efficiency.

\begin{table}[t]
\centering
\setlength{\tabcolsep}{5.2pt}
\begin{tabular}{lccc}
\toprule
\textbf{Model} & \textbf{Params (M)} & \textbf{RTFx} \\
\midrule
$\mathit{VAN}_{H}$ & 71.73 & 1101.63 \\
\midrule
$\mathit{MTL}_{HP}$ & &\\
\, w/ ADC , DII and TIC & 78.71 & 1309.86 \\
\bottomrule
\end{tabular}
\caption{Comparison of model parameters (in millions) and RTFx between the single-task baseline and our proposed method.
 The RTFx represents the mean of 5 independent runs on a single RTX-3090 GPU.
}
\label{tab:params_and_rtf}
\vspace{-10pt}
\end{table}

\section{Conclusion}

This work effectively tackles the inherent challenges of data scarcity and dialectal variability in the development of ASR systems for low-resource languages, with a specific focus on Taiwanese Hakka. 
To achieve this, we proposed and rigorously evaluated a multi-task RNN-T framework that jointly learns multiple linguistic representations and incorporates dialectal information.
By joint training on diverse text representations and mixed-dialect data, the proposed system leverages complementary linguistic and acoustic cues.
This approach reinforces the shared encoder representations of the model and enhances performance across all tasks and dialects, thereby enabling improved generalization in multi-dialect scenarios.
Furthermore, we demonstrated that the selection of the dialect conditioning strategy is critical.
Our proposed Token-Interleaved Conditioning (TIC) strategy, which provides a continuous dialect context, significantly outperformed alternative approaches.
The final model, integrating ADC, DII, and TIC, achieved the best overall performance, confirming that explicitly modeling dialect at both the encoder and decoder levels yields the most robust results.

Notably, despite the incorporation of additional modules, the proposed multi-task learning (MTL) framework introduces only a modest increase in parameters while maintaining the same Real-Time Factor (RTF) as the single-task baseline.
This observation indicates that the approach achieves significant accuracy improvements without compromising inference efficiency.

In summary, this study emphasizes that the joint modeling of diverse linguistic representations, coupled with the integration of dialectal information, offers a practical and efficient approach for developing robust ASR systems in low-resource, multi-dialect settings.

\section{Limitation}

Although this study demonstrates the effectiveness of integrating dialect information and multi-task learning for Taiwanese Hakka ASR, certain limitations remain.

First, low-resource languages like Taiwanese Hakka often lack large-scale corpora.
Future work could address these challenges by leveraging external information sources, such as fine-tuning on pretrained multilingual speech or language models (e.g., wav2vec \cite{baevski2020}, Whisper \cite{radford2022}, or Parakeet-TDT \cite{sekoyan2025}), or incorporating large-scale self-supervised representations to enhance both the acoustic and linguistic modeling capabilities under low-resource conditions.

Another limitation of this study is that the Taiwanese Hakka language encompasses more dialects beyond the three examined here---Sixian, Hailu, and NanSixian---such as Dapu, Raoping, and Zhao'an.
Due to the lack of publicly available speech resources for these dialects, our experiments could not be extended to cover them.
We anticipate that future releases of more comprehensive Taiwanese Hakka datasets will enable broader investigations across additional dialects, allowing for a more complete evaluation of dialectal diversity in Hakka ASR.

\section{Bibliographical References}

\bibliographystyle{lrec2026-natbib}
\bibliography{references.bib}
\end{CJK*}
\end{document}